\documentclass[letterpaper, 10 pt, conference]{ieeeconf}  

\IEEEoverridecommandlockouts                              

\usepackage{graphics} 
\usepackage{cite}
\makeatletter
\@ifundefined{labelindent}{}{\let\labelindent\relax}
\makeatother
\usepackage{enumitem}
\setlist[itemize]{leftmargin=*, labelsep=0.5em, nosep}
\usepackage{graphicx}
\usepackage{dblfloatfix} 
\usepackage{cuted}
\IEEEaftertitletext{\vspace{-3.0\baselineskip}} 
\usepackage{caption}
\usepackage{array}
\usepackage{amsmath,amssymb}
\newtheorem{theorem}{Theorem}
\newtheorem{definition}{Definition}

\newtheorem{lemma}{Lemma}

\usepackage{booktabs,threeparttable,makecell,subcaption,array}
\usepackage[table]{xcolor}
\newcommand{\up}{$\uparrow$}
\newcommand{\down}{$\downarrow$}

\definecolor{HeadGym}{HTML}{E6F2E6}
\definecolor{HeadIsaac}{HTML}{DDEFD3}
\definecolor{HeadGenesis}{HTML}{E5DBF3}
\definecolor{HeadMujoco}{HTML}{F5E6E6}
\definecolor{RowBand}{HTML}{F7F7F7}
\usepackage{hyperref}
\newcommand{\tblsetup}{\setlength{\tabcolsep}{5pt}\renewcommand{\arraystretch}{1.15}}

\renewcommand{\up}{$\uparrow$}
\renewcommand{\down}{$\downarrow$}

\renewcommand{\tblsetup}{%
  \setlength{\tabcolsep}{5pt}%
  \renewcommand{\arraystretch}{1.15}%
}
\author{Zixing Lei$^{1,2, *}$, 
        Zibo Zhou$^{1, *}$, 
        Sheng Yin$^{1, *}$, 
        Yueru Chen$^{1, *}$, 
        Qingyao Xu$^{1}$, 
        Weixin Li$^{2}$, \\
        Yunhong Wang$^{2}$, 
        Bowei Tang$^{1}$, 
        Wei Jing$^{3}$, 
        and Siheng Chen$^{1,3}$
\thanks{$^{*}$Equally contributed.}
\thanks{$^{1}$The authors from Shanghai Jiao Tong University. }%
\thanks{$^{2}$The authors from Zhongguancun Academy.}%
\thanks{$^{3}$The authors from Nerv.ai.}%
}
\begin{document}
\title{\LARGE \bf
PolySim: Bridging the Sim-to-Real Gap for Humanoid Control via Multi-Simulator Dynamics Randomization
}
\maketitle
\begin{strip}
  \centering
  \includegraphics[width=0.9\linewidth]{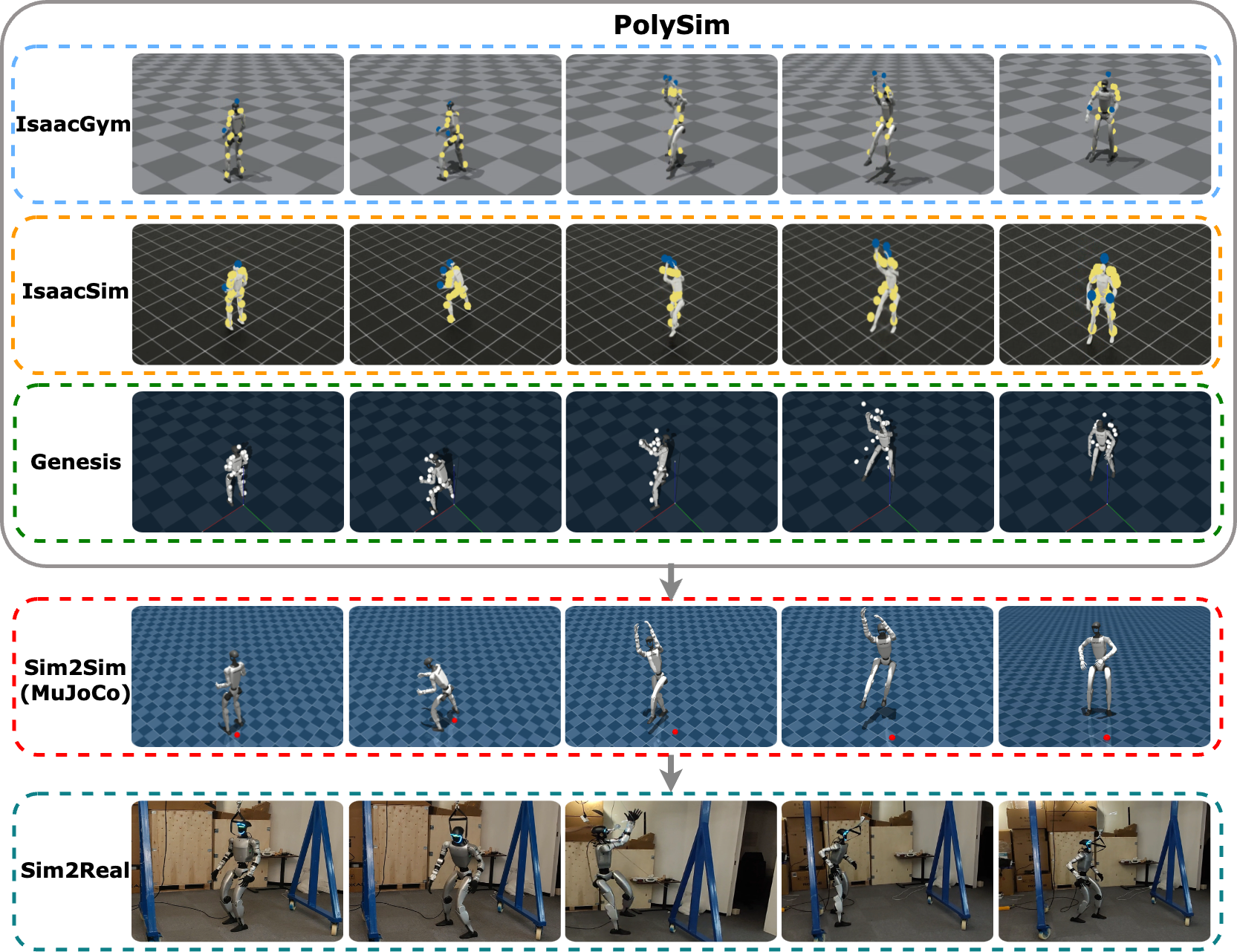}
  \captionof{figure}{\textbf{PolySim}, a parallel training framework, achieves whole-body agility on the Unitree G1 humanoid. Training across diverse simulators reduces motion-tracking error in sim-to-sim transfer and enables zero-shot deployment on the real world.}
  \label{fig:big}
  \vspace{-0\baselineskip}
\end{strip}

\begin{abstract}
Humanoid whole-body control (WBC) policies trained in simulation often suffer from the sim-to-real gap, which fundamentally arises from simulator inductive bias—the inherent assumptions and limitations of any single simulator. These biases lead to nontrivial discrepancies both across simulators and between simulation and the real world. To mitigate the effect of simulator inductive bias, the key idea is to train policies jointly across multiple simulators, encouraging the learned controller to capture dynamics that generalize beyond any single simulator’s assumptions. We thus introduce \textbf{PolySim}, a WBC training platform that integrates multiple heterogeneous simulators. \textbf{PolySim} can launch parallel environments from different engines simultaneously within a single training run, thereby realizing \emph{dynamics-level} domain randomization. Theoretically, we show that PolySim yields a tighter upper bound on simulator inductive bias than single-simulator training. In the experiments, PolySim substantially reduces motion-tracking error in sim-to-sim evaluations; for example, on MuJoCo, it improves execution success by \textbf{52.8\%} over an IsaacSim baseline. PolySim further enables \emph{zero-shot} deployment on a real Unitree~G1 without additional fine-tuning, showing effective transfer from simulation to the real world.  The code is available at \url{https://github.com/EmboMaster/PolySim}

\end{abstract}

\vspace{-2mm}
\section{INTRODUCTION}
Thanks to rapid progress in reinforcement learning (RL) and high-fidelity simulation, humanoid whole-body control (WBC)\cite{yuan2025behavior} has advanced markedly in recent years. Humanoid control is inherently high-dimensional and contact-rich, making the collection of sufficiently diverse real-world interaction data expensive and challenging. Consequently, simulator-based training provides a practical and scalable alternative. Simulation-based trajectory sampling circumvents the safety and efficiency limitations of real-world exploration, enabling the generation of massive rollouts.

\begin{figure}[t]
\centering
    \includegraphics[width=0.48\textwidth]{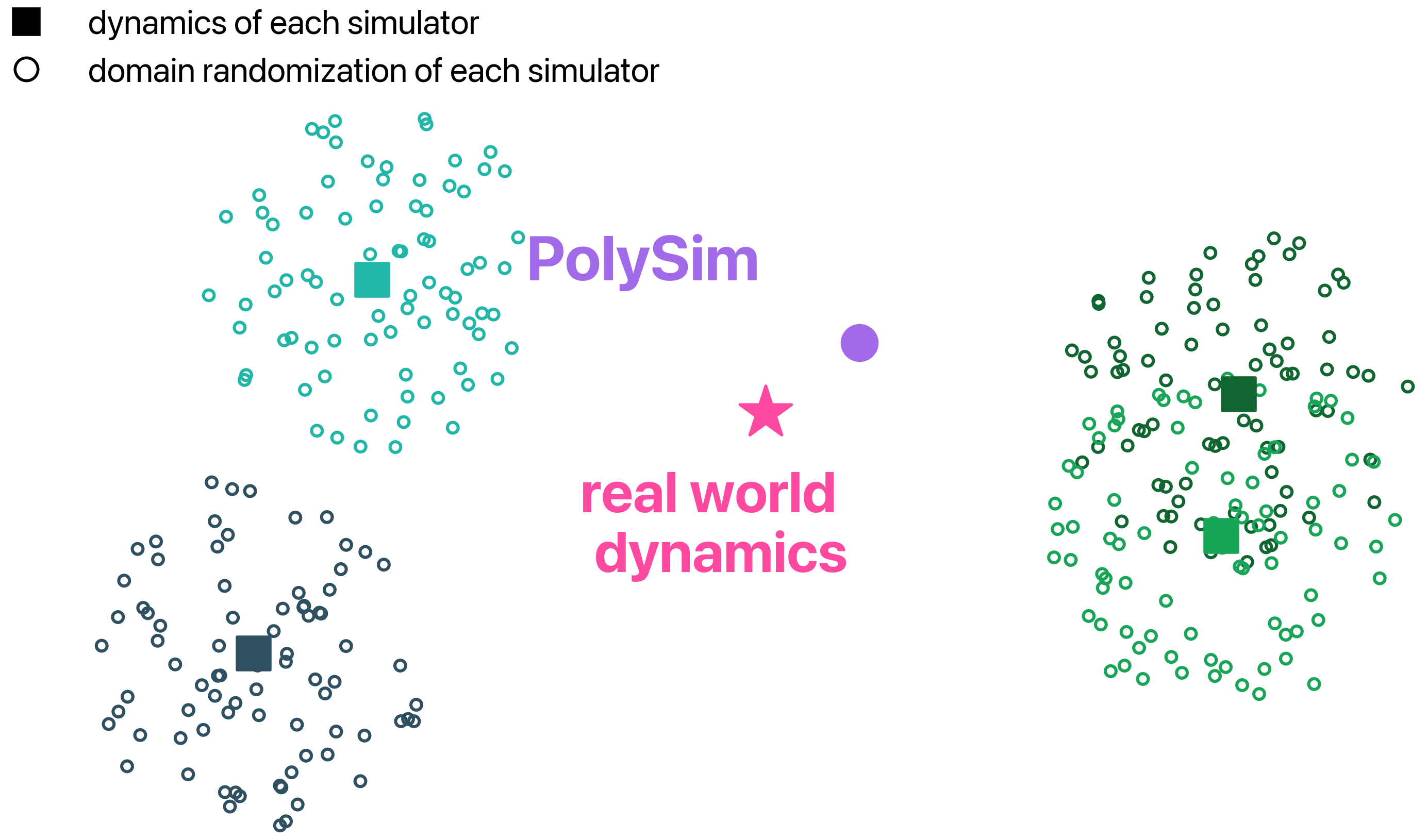}
    \captionof{figure}{Visual illustration of \textbf{PolySim}. The pink star denotes real-world dynamics. Filled squares indicate the nominal transition dynamics of each simulator, representing its inherent inductive bias as an approximation of real-world dynamics. Hollow circles depict domain-randomized variants that perturb parameters but remain centered around their respective simulators. Training against mixtures of simulators (PolySim, purple dot) combines multiple approximations of real-world dynamics, allowing the resulting policy to lie closer to the true dynamics than any single simulator or its parameter-level domain randomization, thereby reducing the notorious sim-to-real gap.
}
    \label{fig:illustrate}
    \vspace{-5mm}
\end{figure}

Despite their utility, simulators provide only approximate models of real-world physics, each encoding its own inherent assumptions. Here we define simulator inductive bias as the inherent modeling assumptions of any simulator that determine its transition dynamics. As shown in Fig.~\ref{fig:illustrate}, every simulator occupies a distinct position in the space of dynamics, while the real world lies elsewhere, with its own intrinsic bias. Policies trained in a particular simulator inevitably inherit its bias, hindering generalization across simulators and to the real world, and causing a significant sim-to-real gap.

To mitigate such a gap, the most common practice is domain randomization\cite{chenunderstanding,tobin2017domain}, which perturbs observations, actions, latent states, or a limited set of physical parameters around a single simulator. This can partially improve generalization by exposing the policy to trajectories slightly beyond the nominal simulator dynamics. Nevertheless, even extensive randomization is structurally limited: the transition model remains that of the chosen simulator, determined by its modeling choices, contact solver, time integrator, and actuator models. Consequently, the randomized rollouts still occupy a narrow neighborhood of that simulator’s dynamics as shown in Fig.~\ref{fig:illustrate}, causing the the sim-to-real gap still substantial.

To address this limitation, our key idea is to train policies jointly across multiple simulators, encouraging the learned controller to capture dynamics that generalize beyond any single simulator’s assumptions. Therefore, we propose \textbf{PolySim}, a WBC training platform that integrates multiple heterogeneous embodied simulators. \textbf{PolySim} enables domain randomization at the level of simulator dynamics, substantially mitigating the effect of any single simulator's inductive bias. Compared with prior work\cite{geng2025roboverse,HumanoidVerseRepo}, \textbf{PolySim} provides three novel features: (i) training–simulation isolation via a client–server architecture that decouples RL training from simulator runtimes and supports flexible distributed execution; (ii) a unified simulator router that performs API translation and resource scheduling, enabling a single training loop to launch parallel environments backed by different simulators; and (iii) GPU pass-through communication that exchanges trajectories and diagnostics directly between training and simulation processes with negligible additional latency. Taken together, these features enable parallel, efficient sampling of diverse trajectories across heterogeneous simulator dynamics during training as shown in Fig~\ref{fig:big}, thereby improving generalization performance.

To validate the effectiveness of \textbf{PolySim}, we analyze its performance both theoretically and empirically.  Theoretically, our analysis shows that the proposed \textbf{PolySim}, which performs dynamics-level randomization, admits a tighter upper bound on simulator inductive bias than approaches that only randomize at the parameter level. Empirically, our extensive experiments show that i) the proposed \textbf{PolySim} significantly reduces motion-tracking error compared to single-simulator baselines in sim-to-sim evaluations. For example, integrating IsaacSim, IsaacGym, and Genesis increases the motion-tracking execution success rate by \textbf{52.8\%} over an IsaacSim-only baseline in MuJoCo; and ii) \textbf{PolySim} enables \emph{zero-shot} deployment on the real Unitree~G1 without additional fine-tuning, reflecting its effectiveness in both simulation and real-world deployment.

\section{RELATED WORK}

\subsection{Learning-Based Methods for Humanoid Control}
Learning-based whole-body control (WBC) for humanoids has rapidly advanced with deep RL~\cite{schulman2017proximal} trained in modern physics simulators~\cite{makoviychuk2021isaac,todorov2012mujoco,mittal2023orbit,mu2021maniskill}. Recent systems demonstrate broad skill coverage: locomotion~\cite{li2019using,li2021reinforcement,liao2025berkeley} on uneven or cluttered terrains~\cite{wang2025beamdojo,gu2024humanoid}, as well as highly dynamic behaviors including stand-up~\cite{he2025learning}, jumping~\cite{li2023robust}, parkour~\cite{long2025learning,zhuang2024humanoid}, and dancing~\cite{zhang2024wococo,ji2024exbody2,cheng2024expressive}. In parallel, physics-based motion imitation continues to push agility and expressivity using human demonstrations and priors~\cite{tessler2024maskedmimic,peng2018deepmimic,xu2025intermimic}. Despite these gains, most controllers are trained in a single simulator. 
\subsection{Sim-to-Real Methods for Robot RL}
Bridging the reality gap typically follows two lines. The first is domain randomization (DR) within one simulator~\cite{tan2018sim,tobin2017domain,mehta2020active,zhou2022domain}, perturbing rendering and physics parameters to improve robustness; this helps but usually preserves the engine’s underlying transition model, leaving model-class mismatch unaddressed. The second is system identification (SysID)~\cite{kozin1986system,aastrom1971system,ljung2010perspectives}, which tunes simulators toward hardware using real data. Offline SysID~\cite{tan2018sim,hwangbo2019learning,khosla1985parameter,yu2019sim,du2021auto,chebotar2019closing} calibrates dynamics before policy learning, while online/adaptive variants~\cite{yu2018policy,peng2020learning,lee2022pi,yu2020learning,yu2017preparing,kumar2021rma,kumar2022adapting} estimate latent properties or adapt policies at deployment. These strategies improve transfer yet often require substantial robot data, can be device-specific, and still inherit the bias of a single engine’s physics. We instead introduce a training-time alternative: parallel exposure to multiple simulators so that the policy optimizes against a mixture of transition models, thereby reducing simulator-induced bias while limiting real-data dependence.
\subsection{Cross-Simulator Frameworks}
\emph{HumanoidVerse} modularizes simulators, tasks, and algorithms to ease switching among engines such as IsaacGym/IsaacSim/Genesis for sim2sim and sim2real studies~\cite{HumanoidVerseRepo}. \emph{RoboVerse} introduces MetaSim, a universal interface that unifies heterogeneous engines and ships with a dataset and standardized benchmarks for IL/RL~\cite{geng2025roboverse}. Both emphasize cross-simulator interoperability, but neither realizes \emph{parallel} cross-environment RL that optimizes a single policy \emph{concurrently} across simulators within one loop.

\section{METHOD}
\begin{figure*}[t]
\centering
    \includegraphics[width=0.85\textwidth]{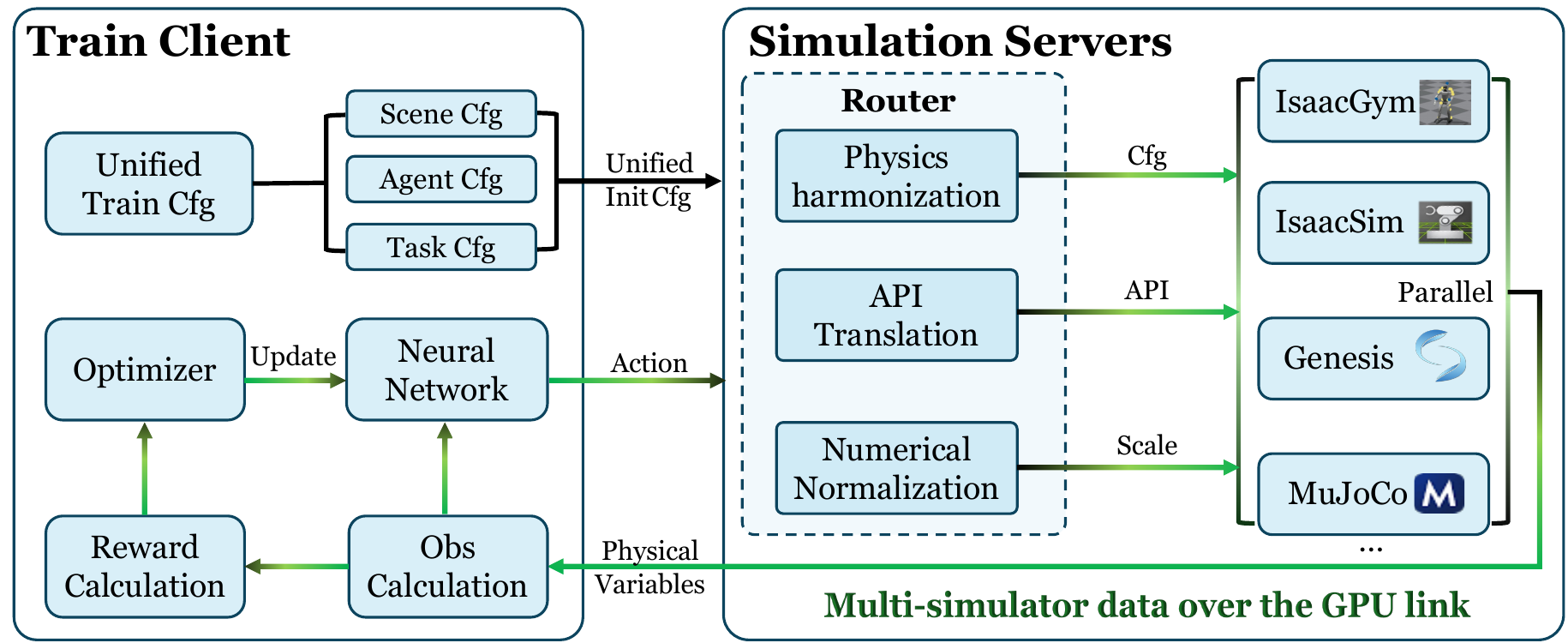}
    \captionof{figure}{System overview of the proposed parallel multi-simulator RL framework (Mode~III). 
\textbf{Left (Training Framework):} a simulator-agnostic RL loop where a unified training configuration (scene/agent/task) drives observation and reward computation; the policy network produces actions and is updated by the optimizer. 
\textbf{Right (Simulation):} heterogeneous engines (IsaacGym/IsaacSim/Genesis/MuJoCo) are virtualized behind a \emph{Simulator Router} that performs \emph{physics harmonization}, \emph{API translation}, and \emph{numerical normalization}. The router maps the unified initialization config to engine-specific settings, dispatches actions, and returns physical variables for observation/reward calculation. 
Green paths indicate GPU-direct links (PyTorch RPC/NCCL over NVLink/PCIe), enabling concurrent rollouts across devices and stable, high-throughput training.}
    \label{fig:sys}
    \vspace{-5mm}
\end{figure*}
To mitigate simulator-induced inductive bias, we operationalize transition-level domain randomization by optimizing a single policy against a mixture of heterogeneous dynamics within each training iteration. \textbf{PolySim} realizes this setting by orchestrating parallel rollouts across multiple simulators while exposing a simulator-agnostic interface to the learner (Fig.~\ref{fig:sys}). Concretely, \textbf{PolySim} comprises three system designs: (i) \textbf{training–simulation isolation}, which decouples the RL learner from per-engine workers to enable distributed execution and failure isolation; (ii) a \textbf{simulator router} that harmonizes physics, translates APIs, and normalizes numerics so that all backends present identical observation–action–reward semantics; and (iii) \textbf{GPU pass-through} communication (RPC over NCCL) that keeps data on device for high-throughput parallel training.
\subsection{Training–Simulation Isolation}
Most existing WBC training stacks adopt a monolithic, centralized design: a single process both instantiates environments and updates the policy. While convenient for development, this conflates simulator and learner runtimes, causing (i) mutual dependency lock in simulation libraries and drivers constrain the training stack and (ii) hard resource contention simulation stepping and backpropagation compete for the same GPU budget.

We therefore execute training and simulation as \emph{isolated processes}: a \emph{TrainClient} and a set of \emph{SimServers}. Each \textit{SimServer} advances its local physics and returns typed payloads $(o, r, d, \mathrm{info})$; the \textit{TrainClient} aggregates trajectories and updates the policy. This separation (1) enables elastic, highly distributed execution across multiple GPUs, (2) preserves toolchain independence so the learner can adopt newer, faster software stacks without being constrained by simulator dependencies, and (3) provides strong fault isolation—failures or stragglers in one engine neither propagate nor block others—allowing heterogeneous simulators to start and progress concurrently without resource interference.

\subsection{Simulator Router — Unified Interface Virtualization}
Heterogeneous simulators differ in scene parameterization, API surfaces, and numerical conventions, which prevents a single learner from receiving consistent inputs. Our simulator router virtualizes multiple simulators as one vectorized environment with invariant semantics, synchronized clocks, and device-resident tensors. As shown in Fig.~\ref{fig:sys}, simulator router establishes a unified standard along three axes: (i) physics harmonization at initialization to construct maximally aligned scenes; (ii) API translation that exposes a common interface and delivers identically shaped observations and reward signals to the TrainClient; and (iii) numerical normalization to ensure that neural-network actions are correctly interpreted by each engine. With these alignments, SR abstracts away backend idiosyncrasies and enables stable, high-throughput parallel rollouts from heterogeneous simulators within a single training loop.

\textbf{Physics Harmonization.}
Simulator router takes a unified physical specification and instantiates per-simulator scenes that are as consistent as possible. It maps friction and contact settings, actuator models and controller gains, gravity and integration step, rigid-body properties, and robot descriptions to the corresponding options of each simulator. Residual mismatches that arise from different contact solvers or integrators are recorded as fixed engine meta-parameters to provide controlled dynamics variation rather than uncontrolled drift.

\textbf{API Translation.}
Simulator router exposes one environment interface to the learner by translating heterogeneous simulator APIs into a common observation–action–reward contract. It aligns state layouts, reference frames, and materializes only the physical quantities required by the policy and the reward. The result is identically shaped tensors with consistent meaning, independent of the underlying engine.

\textbf{Numerical Normalization.}
Simulator router reconciles engine-specific numerical conventions so that neural-network outputs are applied correctly across backends. It performs unit conversion at the interface boundary such as standardizing angular quantities to radians, applies range mapping from a normalized action space such as rescaling $[-1,1]$ actions to per-engine joint limits or position-target ranges, and enforces clipping consistent with actuator and controller limits to guarantee safety and correctness. 

As a result, from the RL framework’s perspective all simulators deliver the same service: a unified interface with identical tensor shapes, semantics, and timing, while simulator router handles backend-specific details and routing.

\subsection{GPU-Direct Communication Pipeline}
To sustain high-throughput parallel rollouts across heterogeneous engines, \textbf{PolySim} employs a RPC pipeline over NCCL. Observations, actions, rewards, and masks are kept as device tensors end to end; cross-process exchanges use NVLink or high-speed PCIe without host staging, which removes CPU round trips and redundant serialization, and preserves bandwidth at RL rollout scales. This capability fully leverage CUDA RPC’s ability to transmit tensors directly from local GPU memory to remote GPU memory and on NCCL’s peer-to-peer transports.

\section{Theoretical Superiority of \textbf{PolySim}}
\textbf{Setup and notation.} Let the real environment be $M^*=(\mathcal S_0,\mathcal A,T_0,R_0,\gamma)$ and each simulator $M_i=(\mathcal S_i,\mathcal A,T_i,R_i,\gamma)$ be a projection via $f_i:\mathcal S_0\!\to\!\mathcal S_i$, where $\mathcal S_0$ is the real state space, $\mathcal A$ the common action space, $T_0(\cdot\mid s,a)$ the transition kernel on $\mathcal S_0$, and $R_0$ a bounded reward; for simulator $i$, $\mathcal S_i$ is its state space with transition kernel $T_i$ and reward $R_i$.
Given a measurable section $g_i:\mathcal S_i\!\to\!\mathcal S_0$ satisfying $f_i\!\circ g_i=\mathrm{id}_{\mathcal S_i}$, define the lifted kernel
\[
\tilde T_i(\cdot\mid s,a)\;:=\;(g_i)_{\#}\,T_i(\cdot\mid f_i(s),a),
\]
so that all kernels act on the common space $\mathcal S_0$. For any policy $\pi$, let
$\eta_{T,R}(\pi):=\mathbb{E}[\sum_{t\ge0}\gamma^t R(s_t,a_t)]$
denote the discounted return under $(T,R)$. On $(\mathcal S_0,d)$, let $W_1$ denote the \emph{1-Wasserstein distance induced by $d$}. Assume that for any $\pi$, the value function $V^{\pi,0}$ is $L_V$-Lipschitz on $(\mathcal S_0,d)$.

\begin{definition}[Sim-to-real gap of a simulator]
Let $P_i$ be a simulator with $\tilde{T_i}$ the corresponding lifted kernel. Then, the sim-to-real gap of $P_i$ under a policy class $\Pi$ is,
\[
G_{\mathrm{S2R}}(P_i;\Pi)
~\triangleq~
\sup_{\pi\in\Pi}\big|
\eta_{\tilde T_i}(\pi) - \eta_{T_0}(\pi) 
\big|.
\]
\end{definition}


\begin{lemma}[Upper-bound of the sim-to-real gap]\label{thm:single}
The sim-to-real gap under a policy class $\Pi$ is upper-bounded as
\[
G_{\mathrm{S2R}}(P_i;\Pi)
~\le~
\frac{\gamma}{1-\gamma}\,L_V\,\Delta_i,
\]
where $\gamma\in(0,1)$ is the discount factor, $L_V$ is the Lipschitz constant of $V^{\pi,0}$ and
$\Delta_i:=\sup_{s,a}W_1\!\big(T_0(\cdot|s,a),\tilde T_i(\cdot|s,a)\big)$
\end{lemma}
\begin{definition}[Sim-to-real gap of \textbf{PolySim}]\label{def:poly-gap}
Let $Q$ be \textbf{PolySim}, a mixture of $N$ simulators, whose transition kernel is
\[
T_Q(\cdot\mid s,a)=\;\sum_{i=1}^N w_i\,\tilde T_i(\cdot\mid s,a)\ \in\ \mathrm{conv}(\mathcal K),
\]
where $w\in\mathbf{\Delta}^{N-1}$ be the weight vector (the probability simplex) and $\mathcal K=\{\tilde T_1,\dots,\tilde T_N\}$ be the lifted simulator transition kernels on $\mathcal S^{0}$.  Then, the sim-to-real gap of $Q$ under a policy class $\Pi$ is
\[
G_{\mathrm{S2R}}(Q;\Pi)
~\triangleq~
\sup_{\pi\in\Pi}\,\big|
\eta_{T_{0},R_{0}}(\pi)-\eta_{\tilde T_Q,R_{0}}(\pi)
\big|,
\]
where $\eta_{T,R}(\pi)$ denotes the discounted return under $(T,R)$.
\end{definition}
\begin{theorem}[Superiority of the PolySim S2R Gap]
\label{thm:polysim-gap}
\textbf{PolySim} achieves a lower sim-to-real gap than any single simulator.
Given the lifted simulator kernels $\mathcal K=\{P_i\}_{i=1}^N$ on $\mathcal S^{0}$ (where $P_i:=\tilde T_i$) and a convex mixture
For any policy class $\Pi$, we have
\[
G_{\mathrm{S2R}}(Q;\Pi)\leq\min_{i} G_{\mathrm{S2R}}(P_i;\Pi).
\]
\end{theorem}
Theorem~\ref{thm:polysim-gap} shows that PolySim is better than any single simulator in terms of the sim-to-real gap. 
The intuition is that mixtures enlarge the feasible set from isolated simulators to their convex hull: when the real dynamics $T_{0}$ are strictly closer (in $W_1$) to $\mathrm{conv}(\mathcal K)$ than to any vertex $P_i$, hence the sim-to-real gap shrinks.

\textbf{Proof:}
 For any policy $\pi$ and any kernel $Q$ on $\mathcal S_0$,
writing the Bellman equations under $(T_0,R_0)$ and $(Q,R_0)$, subtracting, taking
$\|\cdot\|_\infty$, and using the $L_V$-Lipschitzness of $V^{\pi,0}$ together with the
Kantorovich--Rubinstein dual for $W_1$ yields
\[
\big\|V^{\pi,*}-V^{\pi,Q}\big\|_\infty
\;\le\; C\cdot \sup_{s,a} W_1\!\big(T_0(\cdot|s,a),Q(\cdot|s,a)\big).
\]
Evaluating returns under any start-state distribution and taking absolute values gives the
general value-gap bound
\begin{equation}
\big|\eta_{T_0,R_0}(\pi)-\eta_{Q,R_0}(\pi)\big|
\;\le\; C\cdot \sup_{s,a} W_1\!\big(T_0(\cdot|s,a),Q(\cdot|s,a)\big).
\label{eq:gen-gap}
\end{equation}
Taking $\sup_{\pi\in\Pi}$ on both sides, we obtain
\begin{equation}
G_{\mathrm{S2R}}(Q;\Pi)
\;\le\; C\cdot \sup_{s,a} W_1\!\big(T_0(\cdot|s,a),Q(\cdot|s,a)\big).
\label{eq:gen-gap-class}
\end{equation}
Instantiate \eqref{eq:gen-gap-class} with $Q=\tilde T_i$ and denote
$\Delta_i:=\sup_{s,a}W_1\!\big(T_0(\cdot|s,a),\tilde T_i(\cdot|s,a)\big)$ to get
\[
G_{\mathrm{S2R}}(i;\Pi)\;\le\; C\,\Delta_i,
\]
which is exactly the claimed single-simulator upper bound.
Let $\varepsilon_{\mathrm{hull}}:=\inf_{Q\in\mathrm{conv}(\mathcal K)}\sup_{s,a}
W_1\!\big(T_0(\cdot|s,a),Q(\cdot|s,a)\big)$ and take
$Q\in\arg\min_{Q\in\mathrm{conv}(\mathcal K)}\sup_{s,a}W_1(T_0,Q)$.
Applying \eqref{eq:gen-gap-class} to $Q$ gives
$G_{\mathrm{S2R}}(Q;\Pi)\le C\,\varepsilon_{\mathrm{hull}}$,
while applying it to each $\tilde T_i$ gives
$G_{\mathrm{S2R}}(i;\Pi)\le C\,\Delta_i$.
Under the stated assumption
$\varepsilon_{\mathrm{hull}}<\min_i \Delta_i$,
there exists $\delta>0$ s.t. $C\,\varepsilon_{\mathrm{hull}}\le C(\min_i\Delta_i-\delta)$.
By the definition of the supremum and the tightness of the KR-dual bound (standard in communicating MDPs with stationary randomized policies), there exists a sequence
$\{\pi_k\}\subset\Pi$ with
$G_{\mathrm{S2R}}(i^\star;\Pi)\ge C\,\Delta_{i^\star}-o_k(1)$
for some $i^\star\in\arg\min_i\Delta_i$.
Hence, for all sufficiently large $k$,
\begin{align*}
G_{\mathrm{S2R}}&(Q;\Pi)\;<\;C\,\Delta_{i^\star}-o_k(1)
\;\le\\\; &\min_i G_{\mathrm{S2R}}(i;\Pi)
\;\le G_{\mathrm{S2R}}(i^\star;\Pi)\; ,  
\end{align*}
which proves Lemma~\ref{thm:single} and Theorem~\ref{thm:polysim-gap}.
The intuition is geometric: the true world dynamics ($T_0$) and each simulator kernel ($\tilde{T}_i$) are points in a metric space. PolySim's mixed kernel ($Q$) can occupy any point within their convex hull, which we assume is strictly closer to $T_0$ than any single vertex.
The key steps are: (1) Bounding the sim-to-real gap by the 1-Wasserstein distance, which connects the gap to this geometric distance. (2) Applying our assumption that the mixture's minimal distance to reality ($\varepsilon_{\mathrm{hull}}$) is strictly smaller than the best single simulator's distance ($\min_i \Delta_i$), which proves PolySim achieves a tighter theoretical gap.

\begin{table*}[t]
\centering
\caption{Closed-loop motion imitation across test environments. 
Cells with \colorbox{blue!15}{\rule{0pt}{1.6ex}\rule{1.6ex}{0pt}} backgrounds indicate \textit{unseen} settings, where the test simulator was not included in training simulators. 
\textbf{Bold black} denotes the best outcomes under unseen settings.}
\label{tab:cross-sim-split}
\begin{subtable}[t]{\textwidth}
\centering
\tblsetup
\begin{threeparttable}
\begin{tabular}{>{\raggedright\arraybackslash}p{34mm} ccccc ccccc}
\toprule
& \multicolumn{5}{c}{\cellcolor{HeadGym}\textbf{IsaacGym}}
& \multicolumn{5}{c}{\cellcolor{HeadIsaac}\textbf{IsaacSim}} \\
\cmidrule(lr){2-6}\cmidrule(lr){7-11}
\textbf{Training Simulators} & \textbf{Succ \up} & $E_{g\text{-}mpjpe}$ \down & $E_{mpjpe}$ \down & $E_{acc}$ \down & $E_{vel}$ \down
& \textbf{Succ \up} & $E_{g\text{-}mpjpe}$ \down & $E_{mpjpe}$ \down & $E_{acc}$ \down & $E_{vel}$ \down \\
\midrule
Genesis & \cellcolor{blue!15}{0.043} & \cellcolor{blue!15}{201.570} & \cellcolor{blue!15}{97.421} & \cellcolor{blue!15}{4.492} & \cellcolor{blue!15}{13.745} & \cellcolor{blue!15}{0.721} & \cellcolor{blue!15}{\textbf{150.560}} & \cellcolor{blue!15}{\textbf{80.380}} & \cellcolor{blue!15}{6.112} & \cellcolor{blue!15}{\textbf{9.018}} \\
\rowcolor{RowBand} IsaacGym & 1.000 & 140.186 & 49.988 & 4.535 & 7.212 & \cellcolor{blue!15}{0.321} & \cellcolor{blue!15}{236.086} & \cellcolor{blue!15}{97.352} & \cellcolor{blue!15}{\textbf{5.156}} & \cellcolor{blue!15}{12.381} \\
IsaacSim & \cellcolor{blue!15}{0.143} & \cellcolor{blue!15}{189.434} & \cellcolor{blue!15}{83.321} & \cellcolor{blue!15}{4.466} & \cellcolor{blue!15}{13.208} & 0.957 & 121.407 & 44.237 & 3.549 & 6.196 \\
\rowcolor{RowBand} IsaacSim+Genesis & \cellcolor{blue!15}{\textbf{0.400}} & \cellcolor{blue!15}{\textbf{159.611}} & \cellcolor{blue!15}{\textbf{74.142}} & \cellcolor{blue!15}{\textbf{4.400}} & \cellcolor{blue!15}{\textbf{10.608}} & 1.000 & 102.065 & 49.506 & 3.376 & 6.324 \\
IsaacGym+Genesis & 1.000 & 108.971 & 51.898 & 3.624 & 6.380 & \cellcolor{blue!15}{\textbf{0.829}} & \cellcolor{blue!15}{181.457} & \cellcolor{blue!15}{83.016} & \cellcolor{blue!15}{5.939} & \cellcolor{blue!15}{9.988} \\

\rowcolor{RowBand} IsaacSim+IsaacGym & 1.000 & 113.443 & 47.038 & 3.773 & 6.505 & 1.000 & 114.100 & 45.428 & 3.548 & 6.185 \\
IsaacSim+IsaacGym+Genesis & 0.929 & 116.141 & 54.730 & 3.794 & 7.049 & 1.000 & 107.949 & 51.291 & 3.536 & 6.459 \\
\bottomrule
\end{tabular}
\end{threeparttable}
\end{subtable}
\begin{subtable}[t]{\textwidth}
\centering
\tblsetup
\begin{threeparttable}
\begin{tabular}{>{\raggedright\arraybackslash}p{34mm} ccccc ccccc}
\toprule
& \multicolumn{5}{c}{\cellcolor{HeadGenesis}\textbf{Genesis}}
& \multicolumn{5}{c}{\cellcolor{HeadMujoco}\textbf{MuJoCo}} \\
\cmidrule(lr){2-6}\cmidrule(lr){7-11}
\textbf{Training Simulators}
& \textbf{Succ \up} & $E_{g\text{-}mpjpe}$ \down & $E_{mpjpe}$ \down & $E_{acc}$ \down & $E_{vel}$ \down
& \textbf{Succ \up} & $E_{g\text{-}mpjpe}$ \down & $E_{mpjpe}$ \down & $E_{acc}$ \down & $E_{vel}$ \down \\
\midrule
Genesis & 1.000 & 90.545 & 61.549 & 2.974 & 7.495 & \cellcolor{blue!15}{0.121} & \cellcolor{blue!15}{175.671} & \cellcolor{blue!15}{87.237} & \cellcolor{blue!15}{3.888} & \cellcolor{blue!15}{11.088} \\
\rowcolor{RowBand} IsaacGym & \cellcolor{blue!15}{0.229} & \cellcolor{blue!15}{105.378} & \cellcolor{blue!15}{72.596} & \cellcolor{blue!15}{4.661} & \cellcolor{blue!15}{10.136} & \cellcolor{blue!15}{0.500} & \cellcolor{blue!15}{214.159} & \cellcolor{blue!15}{71.286} & \cellcolor{blue!15}{6.191} & \cellcolor{blue!15}{11.910} \\
IsaacSim & \cellcolor{blue!15}{0.600} & \cellcolor{blue!15}{98.852} & \cellcolor{blue!15}{57.019} & \cellcolor{blue!15}{\textbf{3.731}} & \cellcolor{blue!15}{\textbf{8.348}} & \cellcolor{blue!15}{0.036} & \cellcolor{blue!15}{174.838} & \cellcolor{blue!15}{80.567} & \cellcolor{blue!15}{\textbf{{3.361}}} & \cellcolor{blue!15}{10.911} \\
\rowcolor{RowBand} IsaacSim+Genesis & 1.000 & 97.953 & 62.186 & 3.882 & 8.033 & \cellcolor{blue!15}{0.100} & \cellcolor{blue!15}{174.846} & \cellcolor{blue!15}{78.886} & \cellcolor{blue!15}{3.650} & \cellcolor{blue!15}{11.094} \\
IsaacGym+Genesis & 1.000 & 96.060 & 60.939 & 2.990 & 6.617 & \cellcolor{blue!15}{0.400} & \cellcolor{blue!15}{169.467} & \cellcolor{blue!15}{74.854} & \cellcolor{blue!15}{4.922} & \cellcolor{blue!15}{10.214} \\
\rowcolor{RowBand} IsaacSim+IsaacGym & \cellcolor{blue!15}{\textbf{0.943}} & \cellcolor{blue!15}{\textbf{87.784}} & \cellcolor{blue!15}{\textbf{55.801}} & \cellcolor{blue!15}{4.275} & \cellcolor{blue!15}{9.826} & \cellcolor{blue!15}{0.429} & \cellcolor{blue!15}{174.978} & \cellcolor{blue!15}{71.628} & \cellcolor{blue!15}{4.690} & \cellcolor{blue!15}{10.864} \\
IsaacSim+IsaacGym+Genesis & 1.000 & 97.728 & 64.879 & 3.876 & 8.437 & \cellcolor{blue!15}{\textbf{{0.564}}} & \cellcolor{blue!15}{\textbf{{151.924}}} & \cellcolor{blue!15}{\textbf{{65.594}}} & \cellcolor{blue!15}{4.308} & \cellcolor{blue!15}{\textbf{9.329}} \\
\bottomrule
\end{tabular}
\end{threeparttable}
\end{subtable}
\label{maintable}
\vspace{-5mm}
\end{table*}
\section{EXPERIMENT}

In this section, we present experiments to evaluate the robustness of PolySim. Our experiments aim to answer the follow three key questions:
\begin{itemize}
    \item Q1: Can PolySim outperform the policy trained on a single simulator to compensate for the dynamics mismatch?
    \item Q2: Can PolySim outperform policies trained sequentially across simulators?
    \item Q3: Does PolySim work for sim-to-real transfer?
\end{itemize}
\subsection{Experiments Setup and Metrics}
\subsubsection{Simulation Experiments}
We train policies on 14 motions of varying difficulty (easy, medium, hard) selected from the ASAP dataset~\cite{he2025asap}, using their settings to train for 10k, 15k, and 20k iterations, respectively. The training is performed in parallel across a diverse set of simulators, including IsaacGym, IsaacSim, and Genesis. To purely evaluate the generalization benefit of PolySim, these policies are then tested via zero-shot transfer to an entirely unseen simulator (MuJoCo) without applying any parameter-level domain randomization.
\subsubsection{Real-World Experiments}
For real-world evaluation, we deploy the trained policies on a Unitree G1 humanoid robot and execute motion sequences with substantial sim-to-real discrepancies to evaluate PolySim’s effectiveness in bridging the gap between simulation and physical systems. Specifically, we selected some challenging motions, which span a wide range of motor skills and showcase PolySim’s capability for agile whole-body motion control in real-world scenarios.
\subsubsection{Metrics}
To evaluate motion imitation and transfer performance, we adopt the following metrics:
Success Rate: An imitation is deemed unsuccessful if at any point, the mean body position error exceeds 0.5 m. $E_{g\text{-}mpjpe}$ (mm): Global body position tracking error. $E_{mpjpe}$ (mm): Root-relative mean per-joint position error. $E_{acc}$ (mm/frame$^2$): Acceleration error. $E_{vel}$ (mm/frame): Root velocity error. All metrics are computed against the reference motion sequences, and mean values are reported across all sequences.
\begin{figure}[t]
\centering
    \includegraphics[width=0.45\textwidth]{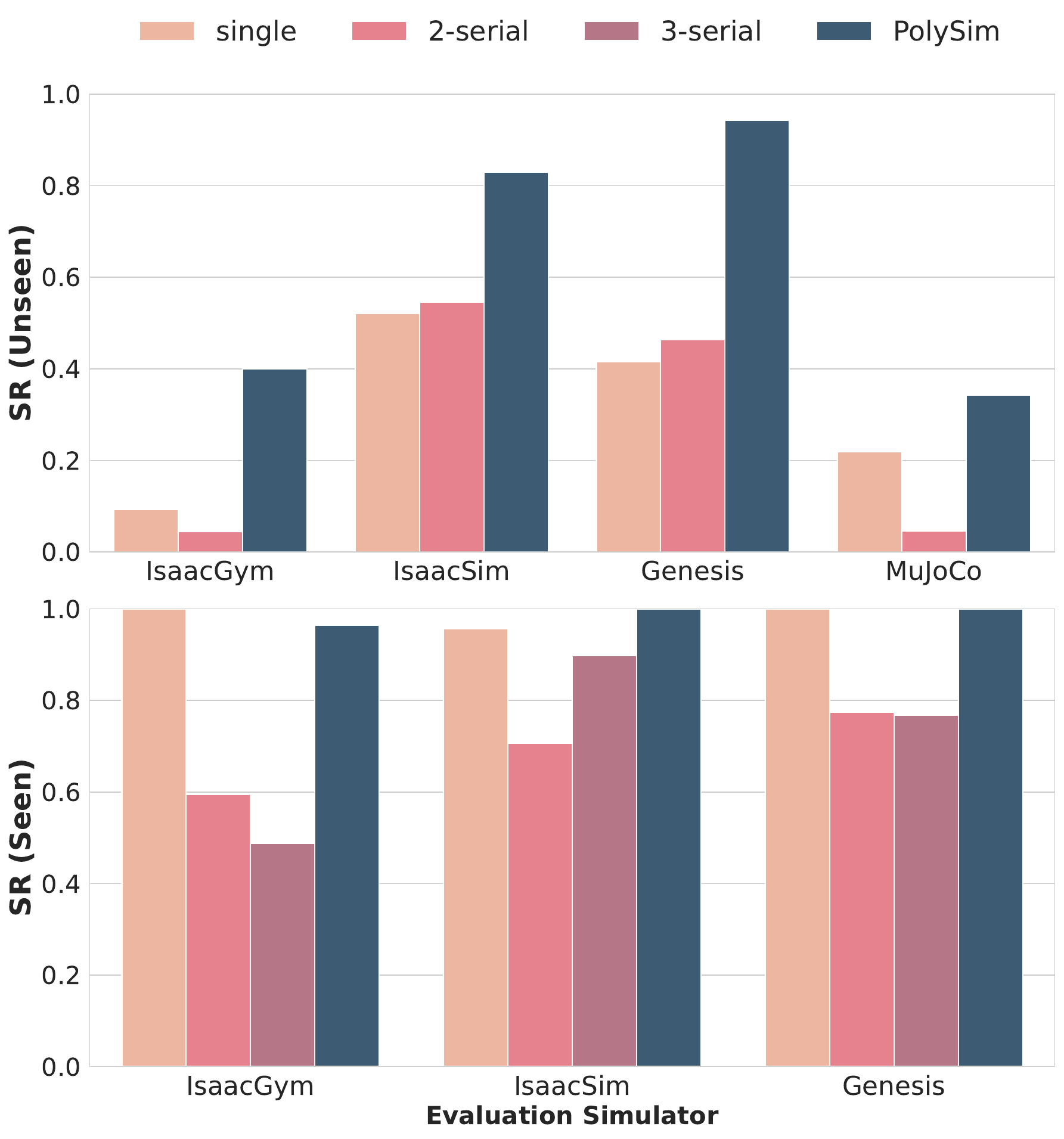}
    \captionof{figure}{Success rate on seen and unseen simulators under different settings. `single' indicates training on a single simulator; `n-serial' indicates sequential training on n simulators.}
    \vspace{-5mm}
    \label{fig:serial}
\end{figure}

\subsection{Baselines}
We compare PolySim against the following baselines: 
\begin{itemize}
    \item \textbf{Single-Domain Training:} Policies are trained independently in each simulator, i.e., IsaacGym, IsaacSim, and Genesis, without heterogeneous domain parallelization.
    \item \textbf{Sequential Multi-Simulator Training:} Policies are trained on multiple simulators sequentially, where two or three simulators are selected in each experiment and executed in different orders for the same number of iterations. Each subsequent simulator inherits the policy and curriculum learning parameters from the previous stage.
\end{itemize}
\subsection{Main Result}
To address \textbf{Q1}, we conduct parallel training with diverse combinations of simulators. We then compare the resulting policies against those trained on a single simulator. Table~\ref{tab:cross-sim-split} reports the performance of \textbf{PolySim} across different test environments. 
The results highlight several key findings:

\subsubsection{Superior generalization to unseen domains}

The most significant advantage of the PolySim strategy lies in its powerful zero-shot generalization capability, the ability to perform tasks in environments not encountered during training. MuJoCo, as an entirely unseen test domain, serves as the ultimate benchmark for evaluating a policy.

In the unseen MuJoCo domain, the Sim+Gym+Genesis configuration, leveraging the most diverse dynamics, achieved the best overall performance. It demonstrated superior robustness—with success rates increasing by $44.3\%$, $6.4\%$, and $52.8\%$ over single-simulator baselines—and superior precision, with the lowest errors across key metrics ($E_{g-mpjpe}$, $E_{mpjpe}$, $E_{vel}$). This performance not only surpasses naive single-simulator training but also more advanced parameter-level DR (Table~\ref{tab:cross-sim-mujoco_dr}). This confirms our central claim: true generalization requires overcoming a single engine's inductive bias, a feat superficial randomization alone cannot achieve.

\subsubsection{Beyond additive effects: amplified gains from multi-simulator parallel training}

The improvement \textbf{PolySim} is not a simple additive effect of single-simulator models, but rather a synergistic gain where joint training produces outcomes far superior to the sum of its parts. For example, in Table~\ref{tab:cross-sim-split} Using IsaacGym as the test environment, we compare policies trained on Genesis, on Sim, and on their combination (Sim+Genesis). Genesis-only training achieves a success rate of 0.043; Sim-only achieves 0.143. Joint training on Sim+Genesis raises the success rate to 0.400 and, across all error metrics, surpasses both single-simulator baselines as well as their additive baseline.

\subsubsection{PolySim enhances in-distribution robustness and precision}
For instance, using \textbf{IsaacGym} as the test environment, we compare training solely on \textbf{IsaacGym} with a mixed setting (\textbf{Gym+Genesis}). Training on Gym alone already achieves a perfect success rate of, but the associated error metrics remain relatively high. In contrast, the Gym+Genesis policy not only preserves the perfect success rate but also substantially reduces these errors respectively, representing the best overall performance in this environment. This result highlights the regularization effect of heterogeneous training: by incorporating diverse dynamics from an additional simulator, the policy avoids overfitting to the simulator inductive bias and instead learns smoother, more generalizable control strategies. Consequently, it achieves higher precision and robustness even when evaluated back on the source domain.
\subsection{Parallel vs. Sequential Multi-Simulator Training}
\begin{table}[b]
\vspace{-3mm}
\caption{Iteration times for PolySim compared to the slowest simulator in each recipe. Values in parentheses indicate the additional cost relative to the slowest simulator.}
\centering
\begin{tabular}{cccc}
\toprule
\thead{Slowest\\Simulator}         & \thead{Iteration\\Times(s)}        & \thead{PolySim\\Recipe}       & \thead{Iteration\\Time(s)} \\ \hline
IsaacGym     & 2.1 & IsaacGym+IsaacSim         & 2.3(+0.2)\\
Genesis & 4.7 & IsaacGym+Genesis     & 4.9(+0.2)\\
Genesis & 4.7 & IsaacSim+Genesis     & 4.8(+0.1)\\
Genesis & 4.7 & IsaacGym+IsaacSim+Genesis & 5.0(+0.3)\\ 
\bottomrule
\end{tabular}
\label{tab:iteration_time}
\vspace{-6mm}
\end{table}

\begin{figure*}[!t]
    \centering
    \includegraphics[width=0.9\linewidth]{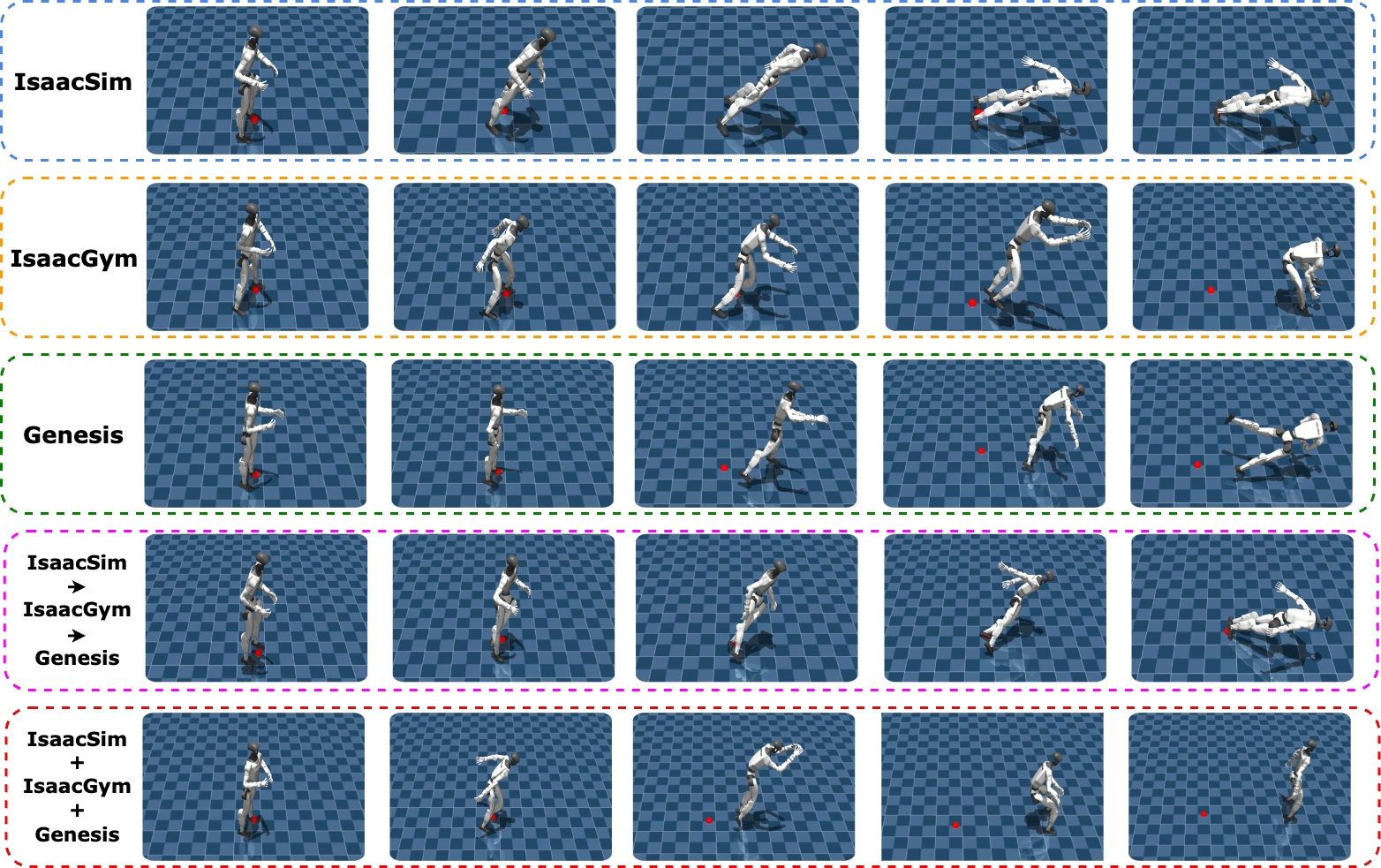}
    \caption{Visualization of sim-to-sim performance in MuJoCo across different training methods. The five panels show the humanoid performing a forward jumping motion under single-simulator, sequential multi-simulator, and PolySim multi-simulator training. Only the PolySim policy trained in parallel across simulators successfully executes the motion in MuJoCo.}
    \vspace{-5mm}
    \label{fig:seqvsmul}
\end{figure*}
To answer \textbf{Q2}, we employ the parallel training framework of PolySim and compare it against policies obtained from sequential training with different simulator permutations, evaluated across five distinct motion tasks. Importantly, all comparisons are conducted under identical RL algorithms and training settings across simulators to ensure fairness.
As shown in Fig \ref{fig:serial}, the results highlight these findings:
\subsubsection{Limited improvement from sequential multi-simulator training}: For unseen evaluation simulators, success rates show only marginal gains compared to single-simulator training—3.5\% and 4.9\% improvements on IsaacSim and Genesis, respectively—while performance on IsaacGym and MuJoCo even decreases. For seen evaluation simulators, sequential training across multiple simulators can actually degrade performance. In particular, 3-serial training performs worse than 2-serial by 10.7\% and 0.6\% on IsaacGym and Genesis, respectively. This anomalous behavior is described as catastrophic forgetting issue and can be observed in Fig~\ref{fig:seqvsmul} also. Our \textbf{PolySim} wtih fully parallel training design significantly improved performance.
\subsection{Experimental efficiency analysis}
TABLE~\ref{tab:iteration_time} reports iteration times for PolySim compared to the slowest simulator in each recipe. The results show that PolySim’s iteration time remains consistently close to that of the slowest simulator, with only a marginal overhead of +0.1–0.3 s. This demonstrates that PolySim achieves efficient multi-simulator training: it leverages diverse physics engines in parallel while incurring negligible additional computation and communication cost.
\subsection{Sim to real deployment} 
To answer \textbf{Q3}, we deploy policies trained with PolySim—using multiple simulator combinations augmented with parameter-based DR—directly onto the Unitree G1 robot. In all cases, the policies achieve successful zero-shot deployment. Detailed motion demonstrations will be provided in our supplementary video materials.
\begin{table}[ht]
\caption{Motion tracking performance of Kobe under different training environments, evaluated on unseen environments.  Here, DR denotes parameter-based Domain Randomization in ASAP~\cite{he2025asap}.}
\centering
\small
\resizebox{\linewidth}{!}{%

\setlength{\tabcolsep}{1mm}{
\begin{tabular}{l|cc|cc}
\toprule
& \multicolumn{2}{c|}{Genesis} & \multicolumn{2}{c}{MuJoCo} \\
Training Env. & Succ $\uparrow$ & $E_{g\text{-}mpjpe}$ & Succ $\uparrow$ & $E_{g\text{-}mpjpe}$ \\
\midrule
$\text{IsaacGym}_{\text{DR}}$ & \textbf{1.000} & 163.135 & 0.100 & 295.877 \\
$\text{IsaacSim}_{\text{DR}}$ & \textbf{1.000} & 130.936 & 0.100 & 272.610 \\
IsaacSim+IsaacGym & \textbf{1.000} & \textbf{103.941} & 0.100 & \textbf{178.190} \\
IsaacSim+IsaacGym+Genesis & - & - & \textbf{1.000} & 199.166 \\
\bottomrule
\end{tabular}}
}
\label{tab:cross-sim-mujoco_dr}
\vspace{-3mm}
\end{table}
\vspace{-2mm}
\section{CONCLUSIONS}
\vspace{-1mm}
In this work, we proposed \textbf{PolySim}, a framework that mitigates simulator inductive bias for robust sim-to-real transfer by training policies across heterogeneous simulators. We provide a theoretical proof that PolySim's dynamics-level randomization admits a tighter upper bound on this bias, a finding validated by extensive experiments. Our results demonstrate superior zero-shot generalization to unseen simulators (e.g., MuJoCo) and enhanced in-domain performance, culminating in the successful zero-shot deployment of a policy onto the Unitree G1 humanoid robot. This work establishes \textbf{PolySim} as a scalable pathway for transferring policies to real-world robots; future work will expand it with more simulators and diverse tasks like manipulation.

\bibliographystyle{IEEEtran}
\bibliography{reference}
\addtolength{\textheight}{-12cm}
\end{document}